\documentclass{article}
\usepackage{amsmath}
\usepackage{hyperref}
\usepackage{graphicx}
\usepackage{booktabs}  
\usepackage{geometry}
\usepackage[backend=bibtex,style=numeric]{biblatex}
\usepackage[T1]{fontenc}
\usepackage{float}

\title{Zero-Shot Forecasting Mortality Rates: A Global Study}
\author{
    Gabor Petnehazi \\
    University of Debrecen\\
    \texttt{gabor.petnehazi@science.unideb.hu}
    \and
    Laith Al Shaggah\\
    University of Debrecen\\
    \texttt{laith.alsaggah@inf.unideb.hu}
    \and
    Jozsef Gall\\
    University of Debrecen\\
    \texttt{gall.jozsef@inf.unideb.hu}
    \and
    Bernadett Aradi\\
    University of Debrecen\\
    \texttt{aradi.bernadett@inf.unideb.hu}
}
\date{}

\begin{document}

\maketitle

\begin{abstract}
This study explores the potential of zero-shot time series forecasting, an innovative approach leveraging pre-trained foundation models, to forecast mortality rates without task-specific fine-tuning. We evaluate two state-of-the-art foundation models, TimesFM and CHRONOS, alongside traditional and machine learning-based methods across three forecasting horizons (5, 10, and 20 years) using data from 50 countries and 111 age groups. In our investigations, zero-shot models showed varying results: while CHRONOS delivered competitive shorter-term forecasts, outperforming traditional methods like ARIMA and the Lee-Carter model, TimesFM consistently underperformed. Fine-tuning CHRONOS on mortality data significantly improved long-term accuracy. A Random Forest model, trained on mortality data, achieved the best overall performance. These findings underscore the potential of zero-shot forecasting while highlighting the need for careful model selection and domain-specific adaptation.
\end{abstract}

\section{Introduction}
Mortality rate forecasting is a critical task with far-reaching implications for public health, insurance, and economic planning. Accurately predicting mortality trends helps inform policy decisions, resource allocation, and financial risk management. However, the task is inherently complex due to unpredictable factors like medical advancements and global events such as pandemics. Traditional forecasting methods often struggle to adapt to these dynamics.

Our study applied zero-shot time series forecasting as an emerging alternative to traditional models. We employed two time-series foundation models alongside a range of traditional and machine learning-based approaches to generate forecasts and conduct evaluations across three forecasting horizons. These evaluations were performed for a diverse set of countries and age groups. Our primary objective was to rigorously validate model performance while demonstrating the potential of zero-shot forecasting in the prediction of mortality rates.


\section{Background}
Historically, mortality rate forecasting has relied on statistical models such as the Lee-Carter (LC) model, see \cite{lee1992modeling}. The LC model decomposes mortality rates into age-specific components and a time-varying index. While effective, these models often lack the flexibility to capture complex, nonlinear patterns in mortality data.

Machine learning methods have recently become widespread, both as stand-alone models and as augmentations to traditional models.

Tree-based machine learning models (random forest, gradient boosting) were used to improve stochastic mortality models, see \cite{levantesi2019application}. All machine learning models produced improvements in forecasting performance, random forest delivering the best results. Deep learning models, particularly convolutional neural networks (CNNs) and recurrent neural networks (RNNs) including long short-term memory (LSTM), are also frequently applied (\cite{hainaut2018neural, richman2021neural, petnehazi2019mortality, nigri2019deep, schnurch2022point}).

Incorporating machine learning models into mortality rate forecasting introduces both powerful capabilities and notable challenges. Machine learning models typically require extensive labeled datasets to perform effectively. In case of mortality rates, the amount of data available is not very large, if only because of the annual observations. Some studies address data scarcity by using data from multiple populations to train the models (see, for example, \cite{scognamiglio2022calibrating, corsaro2024quantile}).

Transformers \cite{vaswani2017attention}, a type of neural network architecture designed to handle sequential data using a mechanism called attention to weigh the importance of different parts of the input, have significantly advanced time series forecasting and machine learning in general. A modified Transformer used to predict mortality rates in major countries was found to give more accurate predictions than the LC model or classic neural networks \cite{wang2024time}. It was also found that Transformers outperform LSTM, ARIMA and simple exponential smoothing \cite{roshani2022transformer}. A Temporal Fusion Tranformer (TFT) \cite{lim2021temporal} outperformed traditional models and CNNs \cite{makhonza2024mortality}.

Perhaps the most impactful contribution of Transformer-based models to machine learning is that they allow training base models with zero-shot capabilities. In times series forecasting context it means that a model trained on a general forecasting task can be directly applied to new, unseen series without task-specific fine-tuning. This may be especially valuable in fields with small, low-frequency data, such as mortality rate forecasting. The present study investigates the applicability of time series foundation models for predicting mortality rates.

\section{Zero-shot Models}
Advancements in machine learning have been significantly driven by the emergence of foundation models---large, pre-trained neural networks capable of tackling a wide range of tasks with minimal or no task-specific fine-tuning. While large language models (LLMs) like GPT \cite{radford2018improving} have revolutionized natural language processing through their zero-shot and few-shot capabilities, this paradigm is now being extended to other domains, including time series analysis. Zero-shot forecasting, a novel approach in this context, leverages the power of these general-purpose models to predict future trends without prior exposure to task-specific data. By learning patterns from diverse and extensive datasets during pre-training, such models generalize well across varied domains, reducing the need for specialized training.

In this study, we explore the potential of zero-shot forecasting by applying two state-of-the-art zero-shot models to the task of mortality rate prediction. These models are evaluated and compared against various established benchmarks, which are discussed in detail in the next section.

TimesFM \cite{das2023decoder} is a decoder-only time series forecasting model with some time series specific model choices (e.g., patching, \cite{nie2022time}). It makes direct multi-step forecasts for a given output patch length. For longer forecast horizons, this direct multi-step prediction is applied in an autoregressive manner. (The output patches are longer than the input patches so that fewer autoregressive steps are necessary.) TimesFM is available with 200M parameters.

CHRONOS \cite{ansari2024chronos} is a simple framework for pretraining forecasting models with no time-series specific modifications. By quantizing the time series, it allows for applying large language models to time series in the same manner as to texts. CHRONOS was used to train both decoder-only (GPT2, \cite{radford2019language}) and encoder-decoder models (T5, \cite{raffel2020exploring}). Pretrained CHRONOS models are available in different sizes from 8M to 710M parameters.

Both models were pretrained on both real and synthetic data. According to the documentations, mortality rate series were not used during the training.

\section{Benchmark Models}
To evaluate the performance of our proposed zero-shot time series forecasting models, we compare them against a diverse set of benchmark models. These benchmarks span traditional time series forecasting techniques, modern methods, and specialized mortality forecasting models, ensuring a comprehensive assessment.
\begin{itemize}
    \item \textbf{Traditional Time Series Models (Naive Benchmarks)}
    \begin{itemize}
        \item \textbf{ARIMA (AutoRegressive Integrated Moving Average)}: ARIMA models \cite{box2015time} are widely used in time series forecasting, leveraging patterns within historical data to predict future trends. For this study, we are using AutoARIMA \cite{hyndman2008automatic, pmdarima}, an automated approach that optimizes ARIMA parameters by selecting the best-fitting model based on statistical criteria. Separate ARIMA models are used to forecast the time series of all country-age group combinations.
        \item \textbf{Exponential Smoothing (ETS)}: This method applies weighted averages of past observations, with exponentially decreasing weights, to predict future values. The ETS framework (see \cite{hyndman2018forecasting}) includes models like simple exponential smoothing, Holt's linear trend method, and Holt-Winters seasonal method, providing a versatile set of tools for different time series patterns. We apply an Exponential Smoothing model from the statsmodels python package \cite{seabold2010statsmodels} with an additive trend and no seasonality. Separate Exponential Smoothing models are used to forecast each mortality rate time series.
    \end{itemize}
    \item \textbf{Machine Learning-Based Time Series Models}
    \begin{itemize}
        \item \textbf{Long Short-Term Memory (LSTM) Networks}: Long Short-Term Memory (LSTM) \cite{hochreiter1997long} networks are a type of recurrent neural network (RNN) designed to model sequential data and capture long-term dependencies. LSTMs excel at time series forecasting by leveraging their unique architecture, which includes memory cells, input gates, output gates, and forget gates. These components allow the model to retain relevant information over extended time horizons while discarding less useful data. We have trained a network in Keras \cite{chollet2015keras} with a single LSTM layer with 50 units, followed by a 1-unit dense layer. We applied a logarithmic transformation to the mortality rates and forecasted the log-transformed values. To ensure the transformation was well-defined, all original values were constrained to a minimum threshold of $1 \times 10^{-6}$, with any values below this threshold clipped to $1 \times 10^{-6}$. All available min-max scaled log-mortality rate series were used to train a single global forecasting model. The data was fed to the model in batches of 512 16-step sequences, for 100 epochs.
        \item \textbf{Random Forest (RF)}: The autoregressive Random Forest approach adapts Random Forest \cite{breiman2001random}, a machine learning ensemble method, for time series forecasting by incorporating lagged values of the series as predictors. This method combines the predictive power of Random Forest with the autoregressive framework, which models the current value of a series as a function of its past values. Random Forest's capability to handle non-linear relationships, high-dimensional data, and complex interactions makes it a versatile tool for forecasting. We have trained a Random Forest in scikit-learn \cite{scikit-learn} with 100 trees, using the past 16 years' mortality rates as input. As with the LSTM, a single global RF model was trained using all available data.
    \end{itemize}
    \item \textbf{Specific Mortality Rate Forecasting Models}
    \begin{itemize}
        \item \textbf{Lee-Carter Model}: A seminal model for mortality forecasting, the Lee-Carter model decomposes mortality rates into an age-specific component and a time-varying factor that captures overall mortality trends. In the original Lee-Carter model \cite{lee1992modeling}, the trend component, extracted from the singular value decoposition (SVD) of the centered log-mortality matrix of the given country, is forecasted with ARIMA(0,1,0). The Lee-Carter model is applied separately to each country.
        \item \textbf{Lee-Carter Variants}: Several extensions of the original Lee-Carter model have been developed to improve flexibility and accuracy. We are applying alternative time series forecasting methods, such as AutoARIMA and zero-shot forecasting methods, to model the trend component. This approach allows us to assess the impact of different forecasting methods on the trend, providing a more flexible framework for capturing complex long-term mortality patterns in the data. We also extend the model by keeping several trend components from the SVD rather than just one (see \cite{booth2001age, baran2007forecasting}), and forecast all components with AutoARIMA.
    \end{itemize}
\end{itemize}

\section{Data}
For our analysis, we utilized mortality rate data from the Human Mortality Database (mortality.org, \cite{HMD}), which provides comprehensive, high-quality datasets for a wide range of countries. This database is widely recognized for its reliability and consistency in recording mortality rates across different regions and time periods. By including all available countries, we aimed to evaluate the zero-shot forecasting models on the widest possible scale. This broad dataset allows for a robust assessment of the models' ability to generalize across diverse demographic profiles, healthcare systems, and mortality trends. In doing so, we ensure that the proposed method is tested in a globally relevant and data-rich environment.

We downloaded all available 1x1 life tables, that is we worked with with annual mortality data and 111 age groups (0 to 110+ years). We ended up with 50 data files, which we refer to as 50 countries, but which actually represent fewer countries. For example for France data are available for both total population and civilian population. We also have multiple data files for Germany, the United Kingdom, and New Zealand. The countries and time periods from which we have and used data are listed in Table~\ref{tab:country_data}.

\begin{table}[H]
\centering
\caption{Countries and periods from which we use data}
\label{tab:country_data}
\begin{tabular}{llrrr}
\toprule
Country Code & Country Name & First Year & Last Year & Years \\
\midrule
AUS & Australia & 1921 & 2021 & 101 \\
AUT & Austria & 1947 & 2019 & 73 \\
BEL & Belgium & 1841 & 2022 & 177 \\
BGR & Bulgaria & 1947 & 2021 & 75 \\
BLR & Belarus & 1959 & 2018 & 60 \\
CAN & Canada & 1921 & 2022 & 102 \\
CHE & Switzerland & 1876 & 2022 & 147 \\
CHL & Chile & 1992 & 2020 & 29 \\
CZE & Czechia & 1950 & 2021 & 72 \\
DEUTE & East Germany & 1956 & 2020 & 65 \\
DEUTNP & Germany & 1990 & 2020 & 31 \\
DEUTW & West Germany & 1956 & 2020 & 65 \\
DNK & Denmark & 1835 & 2023 & 189 \\
ESP & Spain & 1908 & 2021 & 114 \\
EST & Estonia & 1959 & 2019 & 61 \\
FIN & Finland & 1878 & 2023 & 146 \\
FRACNP & France & 1816 & 2022 & 207 \\
FRATNP & France & 1816 & 2022 & 207 \\
GBRCENW & England and Wales & 1841 & 2021 & 181 \\
GBRTENW & England and Wales & 1841 & 2021 & 181 \\
GBR\_NIR & Northern Ireland & 1922 & 2021 & 100 \\
GBR\_NP & United Kingdom & 1922 & 2021 & 100 \\
GBR\_SCO & Scotland & 1855 & 2021 & 167 \\
GRC & Greece & 1981 & 2019 & 39 \\
HKG & Hong Kong & 1986 & 2020 & 35 \\
HRV & Croatia & 2001 & 2020 & 20 \\
HUN & Hungary & 1950 & 2020 & 71 \\
IRL & Ireland & 1950 & 2020 & 71 \\
ISL & Iceland & 1838 & 2022 & 185 \\
ISR & Israel & 1983 & 2016 & 34 \\
ITA & Italy  & 1872 & 2021 & 150 \\
JPN & Japan & 1947 & 2022 & 76 \\
KOR & Republic of Korea & 2003 & 2020 & 18 \\
LTU & Lithuania & 1959 & 2020 & 62 \\
LUX & Luxembourg & 1960 & 2022 & 63 \\
LVA & Latvia & 1959 & 2019 & 61 \\
NLD & Netherlands & 1850 & 2021 & 172 \\
NOR & Norway & 1846 & 2023 & 178 \\
NZL\_MA & New Zealand -- Maori & 1948 & 2008 & 61 \\
NZL\_NM & New Zealand -- Non-Maori & 1901 & 2008 & 108 \\
NZL\_NP & New Zealand & 1948 & 2021 & 74 \\
POL & Poland & 1958 & 2019 & 62 \\
PRT & Portugal & 1940 & 2022 & 83 \\
RUS & Russia & 1959 & 2014 & 56 \\
SVK & Slovakia  & 1950 & 2019 & 70 \\
SVN & Slovenia & 1983 & 2019 & 37 \\
SWE & Sweden & 1751 & 2023 & 273 \\
TWN & Taiwan & 1970 & 2021 & 52 \\
UKR & Ukraine & 1959 & 2013 & 55 \\
USA & The United States of America & 1933 & 2022 & 90 \\
\bottomrule
\end{tabular}
\end{table}

\section{Methods and Evaluation}
In this study, we evaluated the performance of various mortality rate forecasting methods, including traditional statistical models and cutting-edge zero-shot forecasting models, across a diverse set of 50 countries. Our analysis focused on making forward-looking predictions over 5-year, 10-year, and 20-year horizons. To assess the accuracy of the forecasting models, we compared these predictions against actual mortality data from the most recent corresponding historical periods (i.e., the last 5, 10, or 20 years).

We report results for 13 forecasting models. We used 3 Lee Carter-variants: one that uses AutoARIMA to forecast the trend component (LeeCarterAUTO), one that uses the largest CHRONOS model to forecast the trend component (LeeCarterZERO), and one that uses 3 trend components (LeeCarterMULTI). We used pre-trained CHRONOS models with 8M parameters (CHRONOSTiny), 46M parameters (CHRONOSSmall), and 710M parameters (CHRONOSLarge).

For pretraining, we used historical data spanning all countries, excluding the last 20 years. As the training and validation periods differ to some degree for different countries, there may be partial overlap in time between these periods. This could introduce a small degree of information leakage, which is a limitation specific to the mortality-pretrained models: LSTM, RF, and CHRONOSSmallFinetuned. CHRONOSSmallFinetuned is a pretrained small (46M) CHRONOS model that we fine-tuned on the same data we used to train the Random Forest and the Long short-term memory models from scratch.

As a large number of countries and age groups are included in the study, we also report grouped results. The 111 age groups have been categorized into 5 groups: Child ($< 12$ years), Adolescent ($12$–$17$), Adult ($18$–$34$), Middle-aged ($35$–$59$), Senior ($> 59$). The 50 countries have been categorized into 4 groups according to the World Bank classification based on GNI per capita: Low Income ($< \$1{,}146$), Lower-Middle Income ($1{,}146$–$4{,}515$), Higher-Middle Income ($4{,}516$–$14{,}005$), High Income ($> 14{,}005$). Since there is a large variation in the length of annual mortality rate series available in different countries (ranging from 18 to 273 years), we have divided the countries into 4 equal sized bins based on the length of mortality rate history: Q1 ($< 62$ years), Q2 ($62$–$74$), Q3 ($75$–$147$), and Q4 ($> 147$), and we report error measures by these data length categories too.

To ensure the feasibility and validity of predictions, we applied a clipping threshold of $10^{-6}$ to all forecasted mortality rates, thereby eliminating any non-positive values. This constraint was essential to maintain the integrity of the evaluation.

The performance of the models was assessed using the Symmetric Mean Absolute Percentage Error (SMAPE), a widely recognized metric that is well-suited for forecasts involving mortality rates across different age groups and countries. Since mortality rates vary in magnitude across age groups and populations, SMAPE provides a normalized measure of forecast accuracy by accounting for the scale of both observed and predicted values. This ensures meaningful comparisons across diverse demographic and regional contexts, where mortality rates may differ substantially.

Both statistical and practical significance analyses were conducted to compare the proposed methods with traditional approaches. Pairwise Wilcoxon signed-rank tests were conducted to evaluate statistical differences in SMAPE between forecasting methods. Differences in the median SMAPE were calculated to assess practical significance. A threshold of 5\% was used to determine meaningful differences.

\section{Results}
\subsection{5-Year Horizon}
\begin{figure}[H]
\centering
\includegraphics[width=\textwidth]{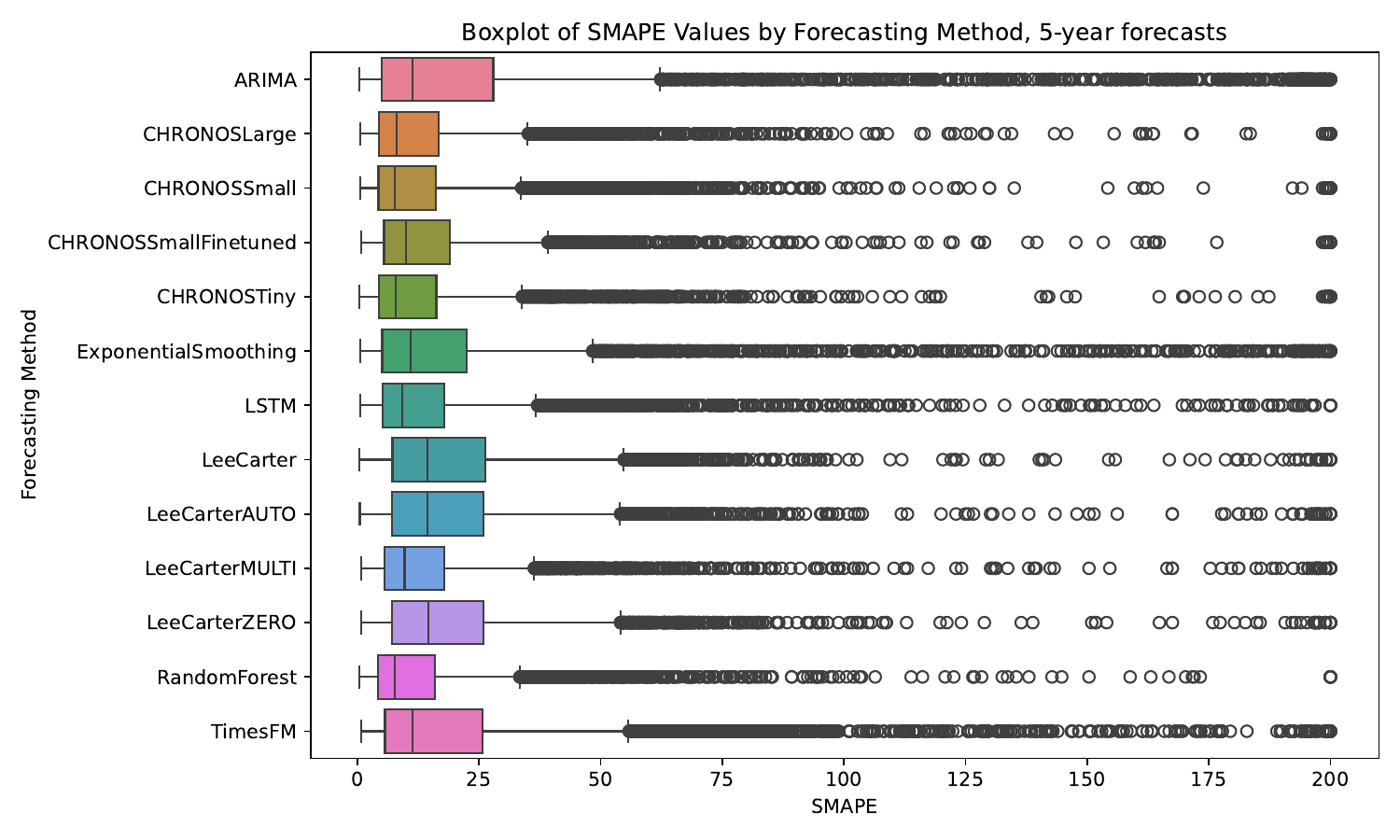}
\caption{Distribution of SMAPE values for each forecasting method, 5-year forecasts}
\label{fig:boxplot_smape_5}
\end{figure}

The mean, median, and standard deviation of SMAPE values for the three methods across 50 countries and 111 age groups are shown in Table~\ref{tab:method_smape_mean_median_std_5}. According to median SMAPE, CHRONOS models of all sizes performed well, outperformed only by the Random Forest model. The TimesFM model is near the bottom of the list in terms of median error, but still above most Lee-Carter models. Figure~\ref{fig:boxplot_smape_5} shows the boxplots of the SMAPE values by forecasting method.

\begin{table}[h!]
\centering
\caption{Mean, median and standard deviation of SMAPE values by forecasting method, 5-year forecasts}
\label{tab:method_smape_mean_median_std_5}
\begin{tabular}{lrrr}
\toprule
Method & \multicolumn{3}{c}{SMAPE} \\
 & mean & median & std \\
\midrule
RandomForest & 13.44 & 7.70 & 17.45 \\
CHRONOSSmall & 14.12 & 7.78 & 19.48 \\
CHRONOSTiny & 14.33 & 7.99 & 19.89 \\
CHRONOSLarge & 14.55 & 8.14 & 19.32 \\
LSTM & 17.05 & 9.22 & 23.83 \\
LeeCarterMULTI & 15.50 & 9.55 & 20.70 \\
CHRONOSSmallFinetuned & 15.43 & 9.90 & 18.07 \\
ExponentialSmoothing & 22.58 & 10.89 & 35.19 \\
TimesFM & 22.31 & 11.35 & 30.27 \\
ARIMA & 30.83 & 11.44 & 48.84 \\
LeeCarterAUTO & 19.71 & 14.00 & 21.48 \\
LeeCarter & 19.78 & 14.12 & 21.35 \\
LeeCarterZERO & 19.64 & 14.27 & 21.04 \\
\bottomrule
\end{tabular}
\end{table}

Table~\ref{tab:method_method_diff_5} shows the model pairs for which the difference in errors was found to be statistically and practically significant. The difference between the errors of the 2 models is considered statistically significant if the null hypothesis of the Wilcoxon signed-rank test (that the differences are symmetric about zero) is rejected at the 5\% significance level. The difference is considered practically significant if the difference between the median errors of the two models is at least 5\% points. Due to the large number of forecasts, the difference was found to be statistically significant in most cases, with practical significance proving to be a stronger constraint. The Lee-Carter model underperforms the generally pretrained CHRONOS models and the RF and LSTM models trained on mortality data.

\begin{table}[h!]
\centering
\caption{Wilcoxon signed-rank test p-values and differences in median errors, 5-year forecasts}
\label{tab:method_method_diff_5}
\begin{tabular}{llrr}
\toprule
Method 1 & Method 2 & Wilcoxon P-value & Difference of Medians \\
\midrule
RandomForest & LeeCarterZERO & 0.00 & -6.57 \\
CHRONOSSmall & LeeCarterZERO & 0.00 & -6.49 \\
RandomForest & LeeCarter & 0.00 & -6.42 \\
CHRONOSSmall & LeeCarter & 0.00 & -6.34 \\
RandomForest & LeeCarterAUTO & 0.00 & -6.30 \\
CHRONOSTiny & LeeCarterZERO & 0.00 & -6.28 \\
CHRONOSSmall & LeeCarterAUTO & 0.00 & -6.22 \\
CHRONOSTiny & LeeCarter & 0.00 & -6.13 \\
CHRONOSLarge & LeeCarterZERO & 0.00 & -6.13 \\
CHRONOSTiny & LeeCarterAUTO & 0.00 & -6.01 \\
CHRONOSLarge & LeeCarter & 0.00 & -5.98 \\
CHRONOSLarge & LeeCarterAUTO & 0.00 & -5.86 \\
LSTM & LeeCarterZERO & 0.00 & -5.05 \\
\bottomrule
\end{tabular}
\end{table}

\begin{figure}[H]
\centering
\includegraphics[width=\textwidth]{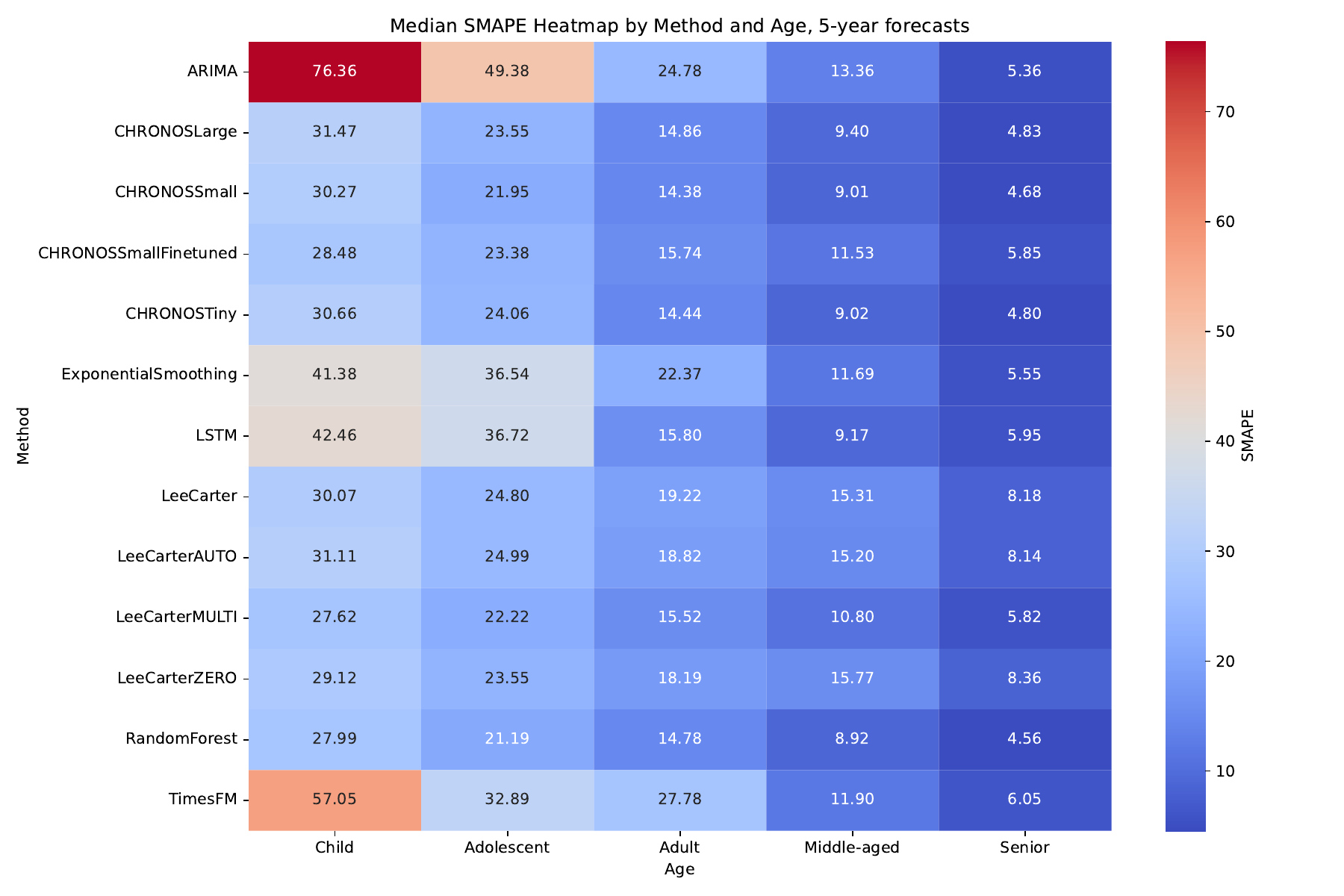}
\caption{Heatmap of median SMAPE by age categories, 5-year forecasts}
\label{fig:method_age_smape_median_5}
\end{figure}

\begin{figure}[H]
\centering
\includegraphics[width=\textwidth]{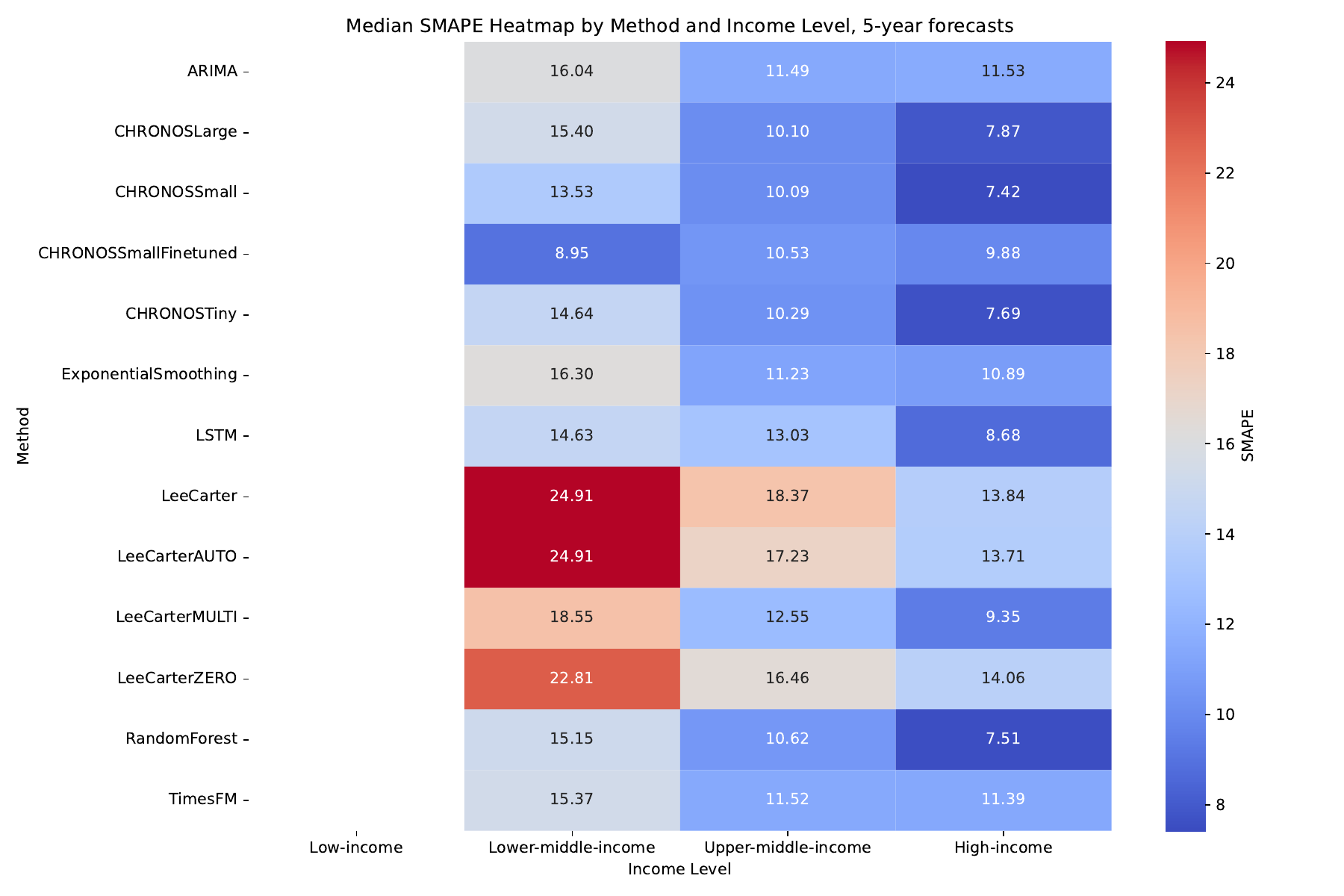}
\caption{Heatmap of median SMAPE by income categories, 5-year forecasts}
\label{fig:method_income_smape_median_5}
\end{figure}

\begin{figure}[H]
\centering
\includegraphics[width=\textwidth]{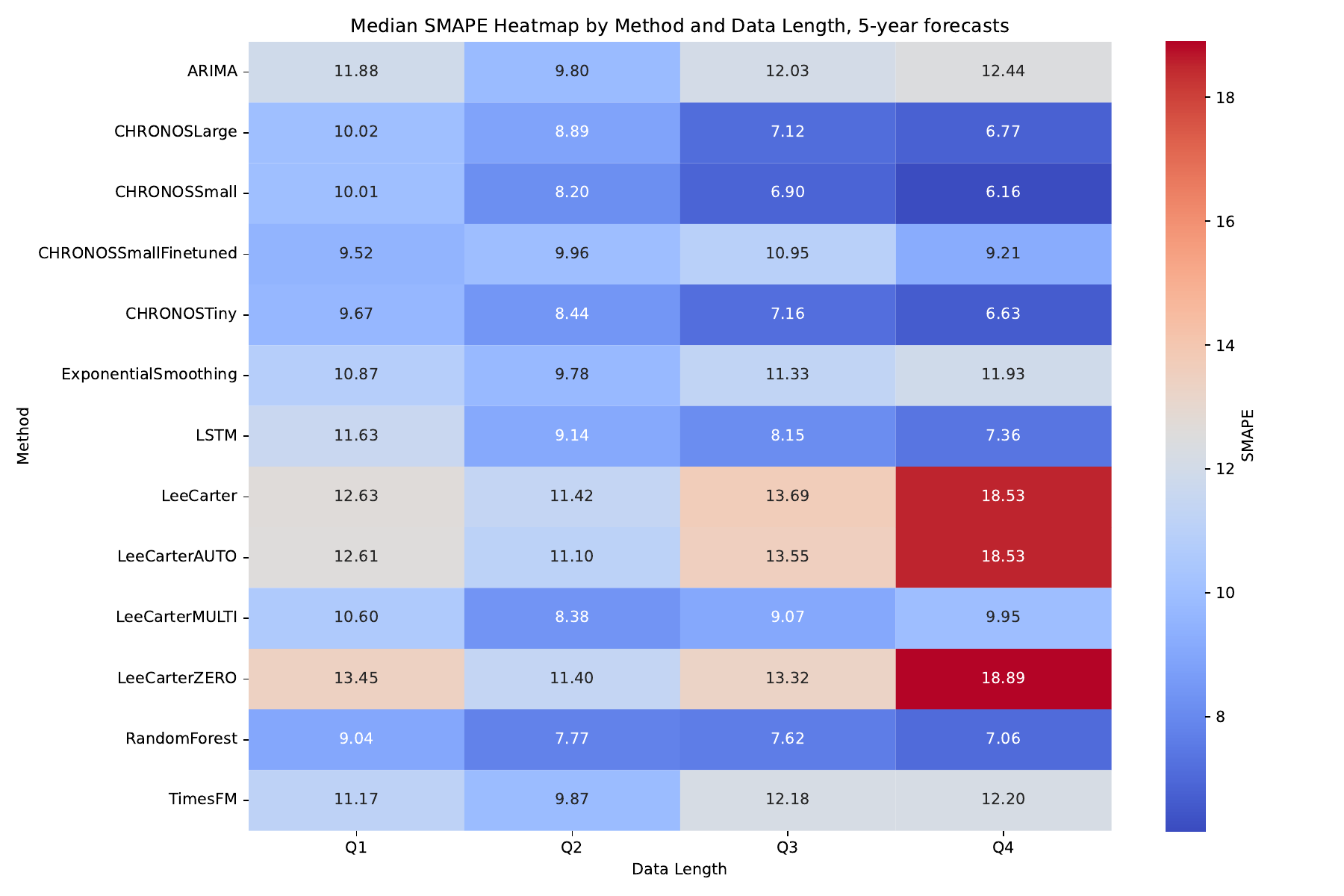}
\caption{Heatmap of median SMAPE by data length categories, 5-year forecasts}
\label{fig:method_length_smape_median_5}
\end{figure}

The median errors by age category, income category and data length category are shown in Figures \ref{fig:method_age_smape_median_5}, \ref{fig:method_income_smape_median_5} and \ref{fig:method_length_smape_median_5}, respectively. In general, the forecasting models seem to produce larger errors for younger ages. Some poorly performing models (e.g. ARIMA, TimesFM) performed comparably at older ages, but produced strikingly large errors in predicting mortality rates at younger ages. Forecasting models generally performed better in predicting mortality rates in higher income countries, and this is true for the zero-shot models as well. The Lee-Carter models performed spectacularly poorly for countries with long mortality rate histories, while the CHRONOS models delivered the best results here.

\subsection{10-Year Horizon}
\begin{figure}[H]
\centering
\includegraphics[width=\textwidth]{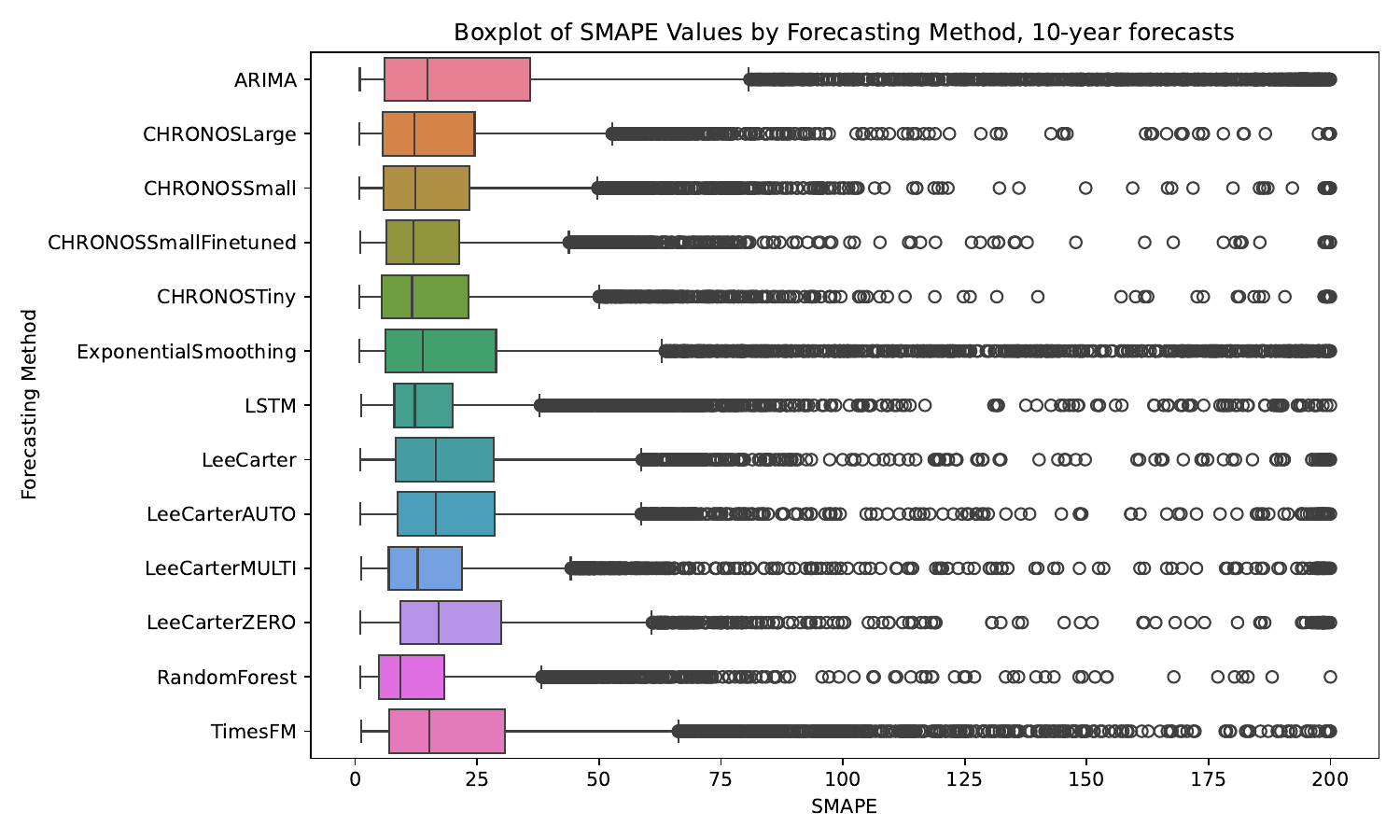}
\caption{Distribution of SMAPE values for each forecasting method, 10-year forecasts}
\label{fig:boxplot_smape_10}
\end{figure}

The mean, median, and standard deviation of SMAPE values for the three methods are shown in Table~\ref{tab:method_smape_mean_median_std_10}. One notable difference is that the fine-tuned CHRONOS model was ranked higher in terms of median error than in the 5-year forecasts. Figure~\ref{fig:boxplot_smape_10} shows the boxplots of the SMAPE values by forecasting method.

\begin{table}[h!]
\centering
\caption{Mean, median and standard deviation of SMAPE values by forecasting method, 10-year forecasts}
\label{tab:method_smape_mean_median_std_10}
\begin{tabular}{lrrr}
\toprule
Method & \multicolumn{3}{c}{SMAPE} \\
 & mean & median & std \\
\midrule
RandomForest & 14.62 & 9.34 & 16.71 \\
CHRONOSTiny & 18.21 & 11.89 & 20.52 \\
CHRONOSSmallFinetuned & 16.83 & 12.07 & 17.49 \\
LSTM & 19.46 & 12.21 & 23.74 \\
CHRONOSLarge & 18.65 & 12.31 & 19.84 \\
CHRONOSSmall & 18.53 & 12.55 & 20.30 \\
LeeCarterMULTI & 18.38 & 13.03 & 21.40 \\
ExponentialSmoothing & 28.28 & 13.90 & 40.08 \\
ARIMA & 36.53 & 15.34 & 51.79 \\
TimesFM & 25.46 & 15.53 & 29.85 \\
LeeCarter & 21.68 & 16.49 & 22.24 \\
LeeCarterAUTO & 22.05 & 16.89 & 22.15 \\
LeeCarterZERO & 22.38 & 17.34 & 21.22 \\
\bottomrule
\end{tabular}
\end{table}

Table~\ref{tab:method_method_diff_10} shows the model pairs for which the difference in errors was found to be statistically and practically significant. We identified fewer practically significant differences here than in the case of the 5-year forecasts.

\begin{table}[h!]
\centering
\caption{Wilcoxon signed-rank test p-values and differences in median errors, 10-year forecasts}
\label{tab:method_method_diff_10}
\begin{tabular}{llrr}
\toprule
Method 1 & Method 2 & Wilcoxon P-value & Difference of Medians \\
\midrule
RandomForest & LeeCarterZERO & 0.00 & -8.00 \\
RandomForest & LeeCarterAUTO & 0.00 & -7.55 \\
RandomForest & LeeCarter & 0.00 & -7.15 \\
RandomForest & TimesFM & 0.00 & -6.19 \\
RandomForest & ARIMA & 0.00 & -5.99 \\
CHRONOSTiny & LeeCarterZERO & 0.00 & -5.45 \\
CHRONOSSmallFinetuned & LeeCarterZERO & 0.00 & -5.26 \\
LSTM & LeeCarterZERO & 0.00 & -5.12 \\
CHRONOSLarge & LeeCarterZERO & 0.00 & -5.03 \\
\bottomrule
\end{tabular}
\end{table}

\begin{figure}[H]
\centering
\includegraphics[width=\textwidth]{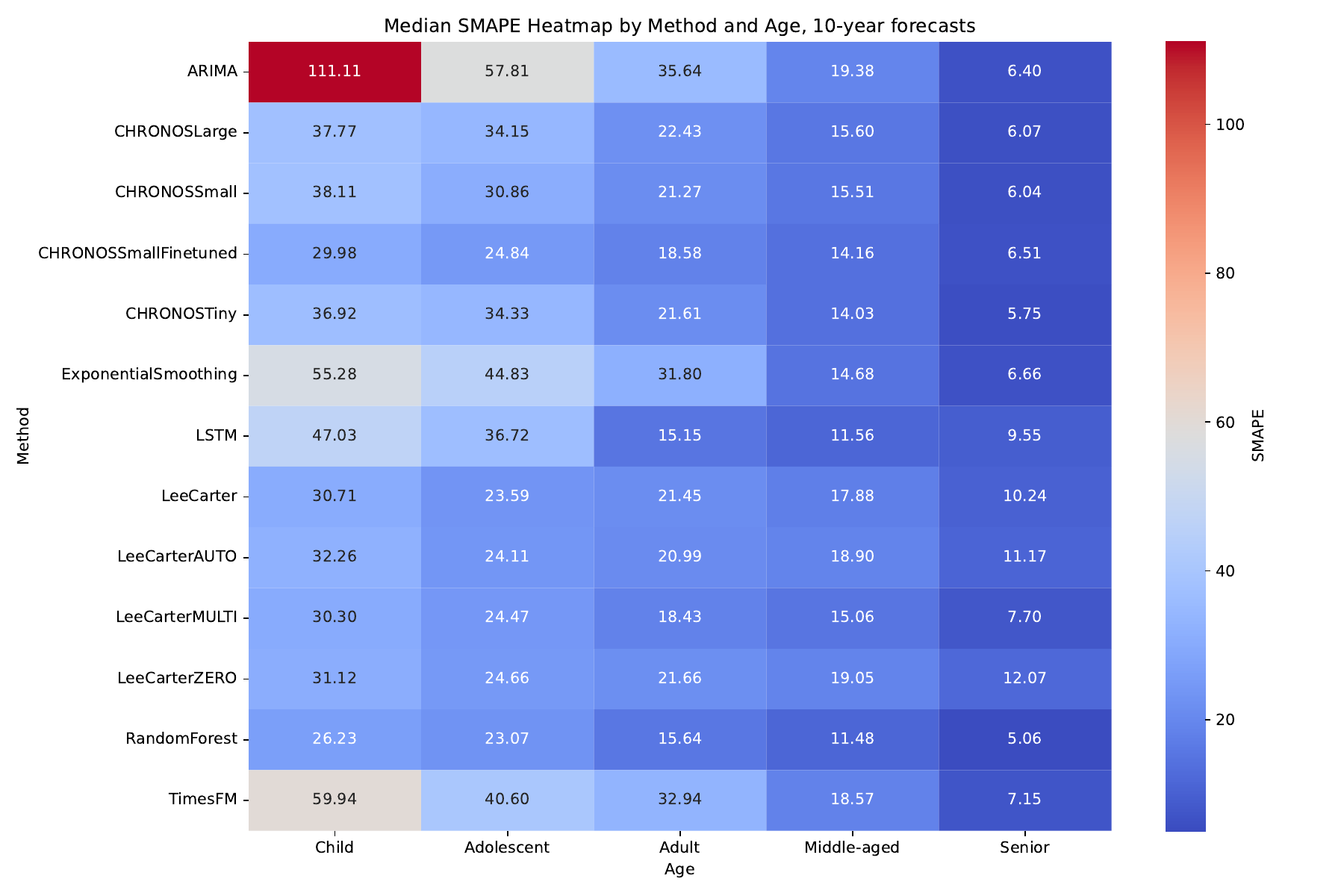}
\caption{Heatmap of median SMAPE by age categories, 10-year forecasts}
\label{fig:method_age_smape_median_10}
\end{figure}

\begin{figure}[H]
\centering
\includegraphics[width=\textwidth]{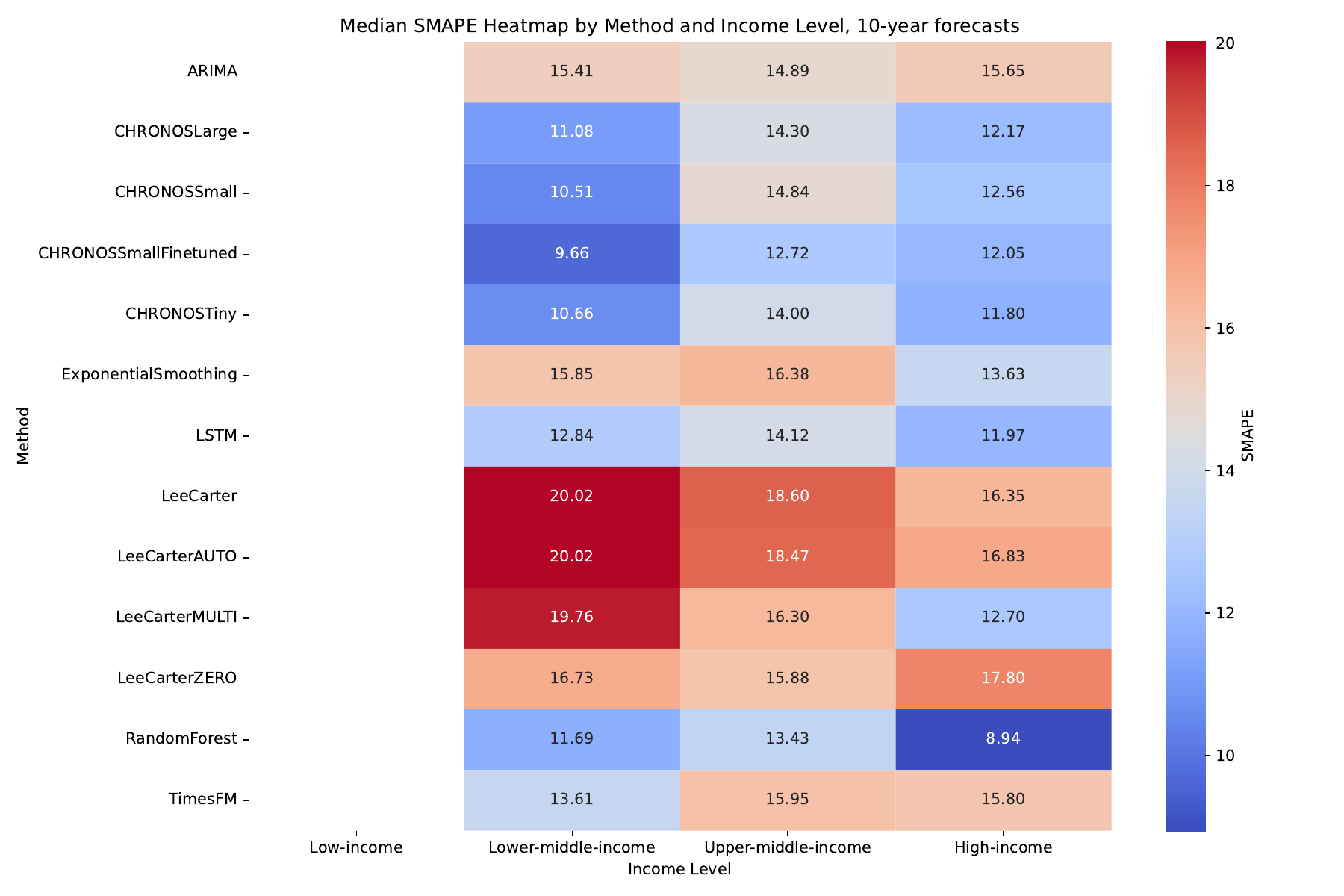}
\caption{Heatmap of median SMAPE by income categories, 10-year forecasts}
\label{fig:method_income_smape_median_10}
\end{figure}

\begin{figure}[H]
\centering
\includegraphics[width=\textwidth]{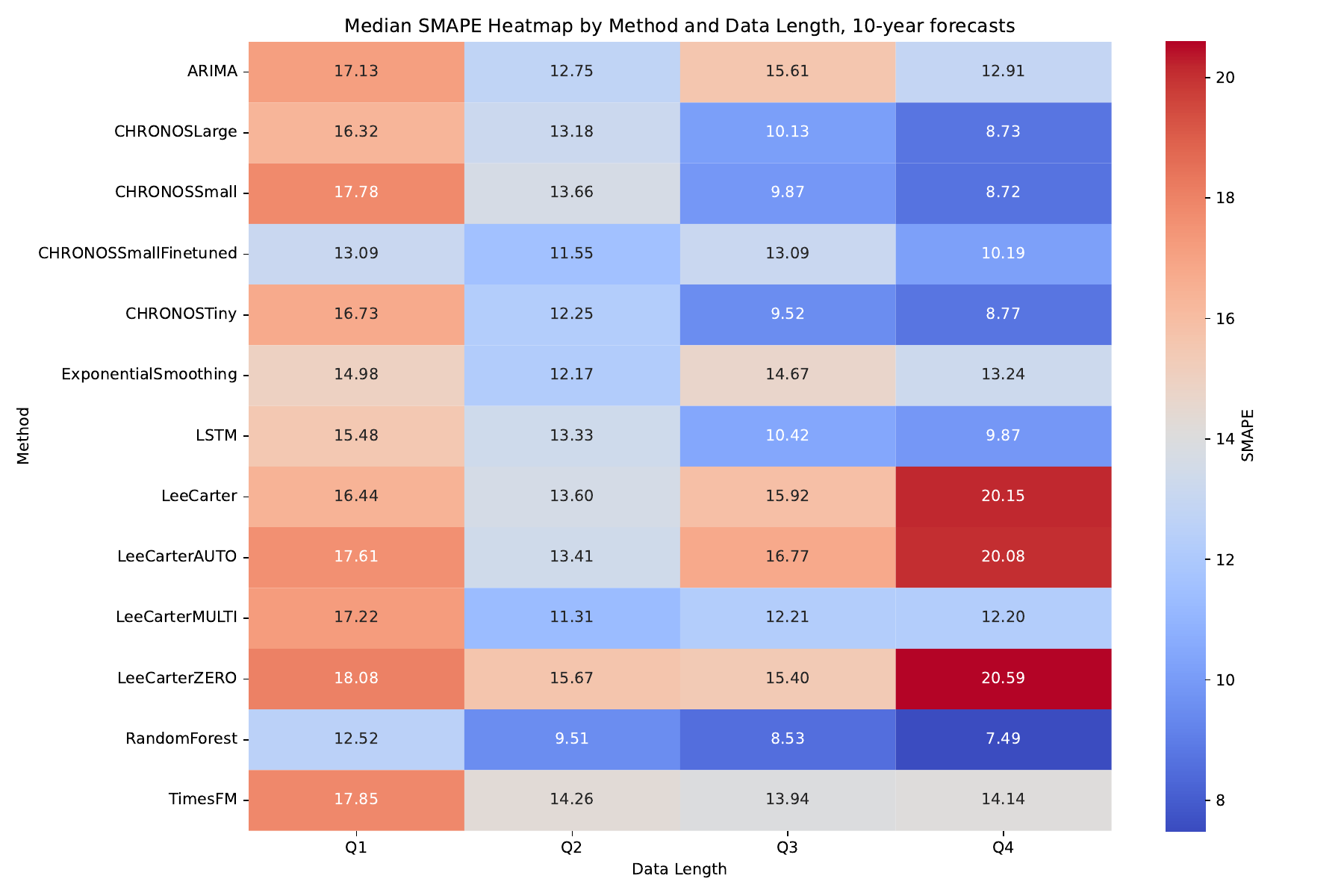}
\caption{Heatmap of median SMAPE by data length categories, 10-year forecasts}
\label{fig:method_length_smape_median_10}
\end{figure}

The median errors by age category, income category and data length category are shown in Figures \ref{fig:method_age_smape_median_10}, \ref{fig:method_income_smape_median_10} and \ref{fig:method_length_smape_median_10}, respectively. Basically similar patterns can be observed as for the 5-year projections. It seems that in this case, the models tend to perform worse for countries with the shortest mortality rate history, but this is not so true for the mortality-fine-tuned CHRONOS model.

\subsection{20-Year Horizon}
\begin{figure}[H]
\centering
\includegraphics[width=\textwidth]{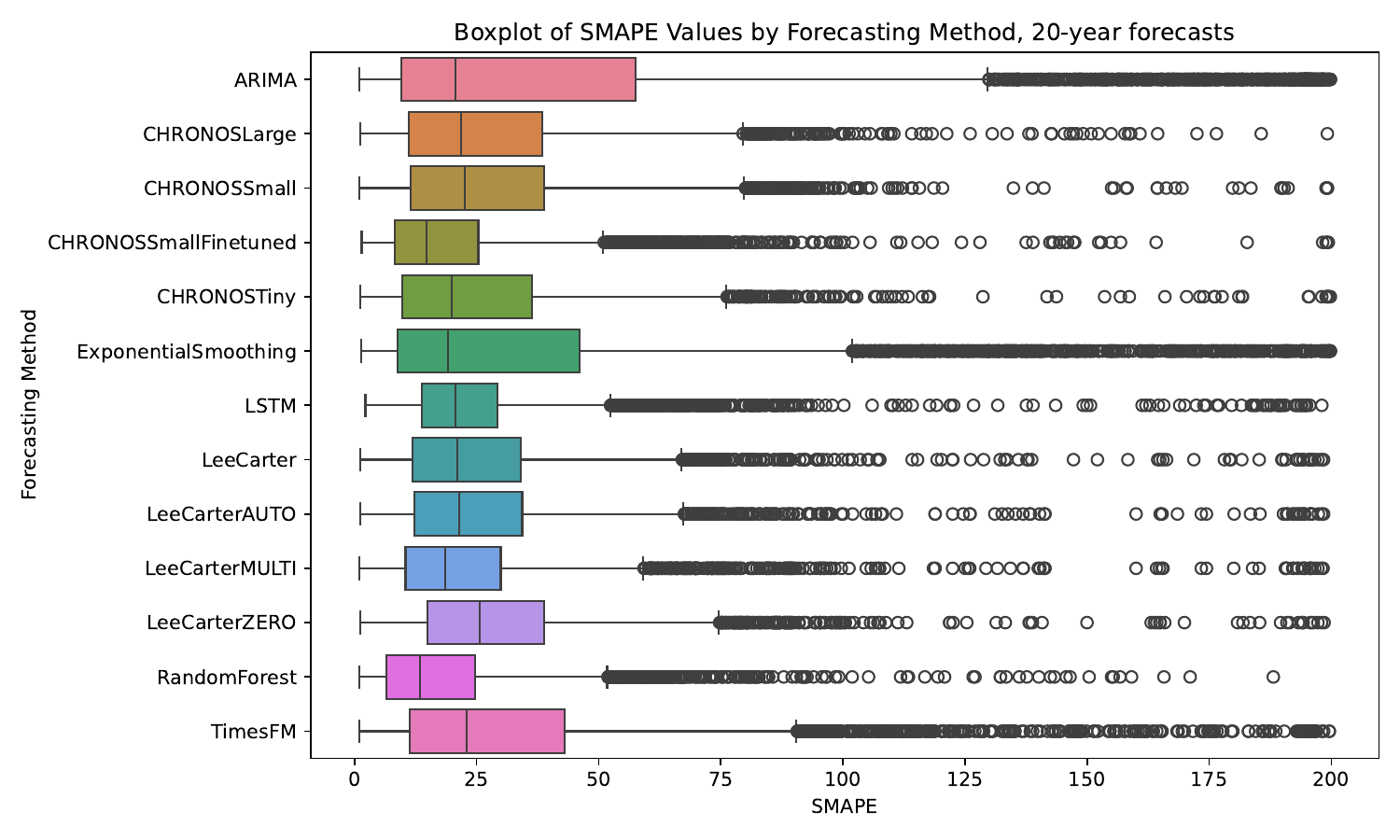}
\caption{Distribution of SMAPE values for each forecasting method, 20-year forecasts}
\label{fig:boxplot_smape_20}
\end{figure}

The mean, median, and standard deviation of SMAPE values for the three methods are shown in Table~\ref{tab:method_smape_mean_median_std_20}. The pre-trained CHRONOS models perform worse in the longest-term prediction, while the CHRONOS model fine-tuned on mortality data has moved up to second place in terms of median error. TimesFM is still an underperformer. Figure~\ref{fig:boxplot_smape_20} shows the boxplots of the SMAPE values by forecasting method.

\begin{table}[h!]
\centering
\caption{Mean, median and standard deviation of SMAPE values by forecasting method, 20-year forecasts}
\label{tab:method_smape_mean_median_std_20}
\begin{tabular}{lrrr}
\toprule
Method & \multicolumn{3}{c}{SMAPE} \\
 & mean & median & std \\
\midrule
RandomForest & 18.63 & 13.36 & 17.98 \\
CHRONOSSmallFinetuned & 19.95 & 14.72 & 18.20 \\
LeeCarterMULTI & 23.03 & 18.47 & 21.14 \\
ExponentialSmoothing & 38.94 & 19.35 & 46.45 \\
CHRONOSTiny & 26.03 & 20.30 & 22.26 \\
LeeCarter & 25.00 & 20.46 & 21.61 \\
LSTM & 25.79 & 20.61 & 22.19 \\
LeeCarterAUTO & 25.33 & 21.05 & 21.40 \\
ARIMA & 45.04 & 21.11 & 54.34 \\
CHRONOSLarge & 27.46 & 22.32 & 21.81 \\
CHRONOSSmall & 27.90 & 22.91 & 22.21 \\
TimesFM & 33.11 & 23.79 & 32.32 \\
LeeCarterZERO & 29.05 & 25.46 & 21.19 \\
\bottomrule
\end{tabular}
\end{table}

Table~\ref{tab:method_method_diff_20} shows the model pairs for which the difference in errors was found to be statistically and practically significant. Now general pre-trained time series forecasting models (both CHRONOS and TimesFM) are significantly outperformed by models trained on mortality data (including the fine-tuned CHRONOS).

\begin{table}[h!]
\centering
\caption{Wilcoxon signed-rank test p-values and differences in median errors, 20-year forecasts}
\label{tab:method_method_diff_20}
\begin{tabular}{llrr}
\toprule
Method 1 & Method 2 & Wilcoxon P-value & Difference of Medians \\
\midrule
RandomForest & LeeCarterZERO & 0.00 & -12.10 \\
CHRONOSSmallFinetuned & LeeCarterZERO & 0.00 & -10.74 \\
RandomForest & TimesFM & 0.00 & -10.43 \\
RandomForest & CHRONOSSmall & 0.00 & -9.55 \\
CHRONOSSmallFinetuned & TimesFM & 0.00 & -9.06 \\
RandomForest & CHRONOSLarge & 0.00 & -8.96 \\
CHRONOSSmallFinetuned & CHRONOSSmall & 0.00 & -8.19 \\
RandomForest & ARIMA & 0.00 & -7.75 \\
RandomForest & LeeCarterAUTO & 0.00 & -7.69 \\
CHRONOSSmallFinetuned & CHRONOSLarge & 0.00 & -7.60 \\
RandomForest & LSTM & 0.00 & -7.25 \\
RandomForest & LeeCarter & 0.00 & -7.10 \\
LeeCarterMULTI & LeeCarterZERO & 0.00 & -6.99 \\
RandomForest & CHRONOSTiny & 0.00 & -6.94 \\
CHRONOSSmallFinetuned & ARIMA & 0.00 & -6.39 \\
CHRONOSSmallFinetuned & LeeCarterAUTO & 0.00 & -6.33 \\
RandomForest & ExponentialSmoothing & 0.00 & -5.99 \\
CHRONOSSmallFinetuned & LSTM & 0.00 & -5.88 \\
CHRONOSSmallFinetuned & LeeCarter & 0.00 & -5.73 \\
CHRONOSSmallFinetuned & CHRONOSTiny & 0.00 & -5.58 \\
LeeCarterMULTI & TimesFM & 0.00 & -5.31 \\
CHRONOSTiny & LeeCarterZERO & 0.00 & -5.16 \\
RandomForest & LeeCarterMULTI & 0.00 & -5.11 \\
LeeCarter & LeeCarterZERO & 0.00 & -5.01 \\
\bottomrule
\end{tabular}
\end{table}

\begin{figure}[H]
\centering
\includegraphics[width=\textwidth]{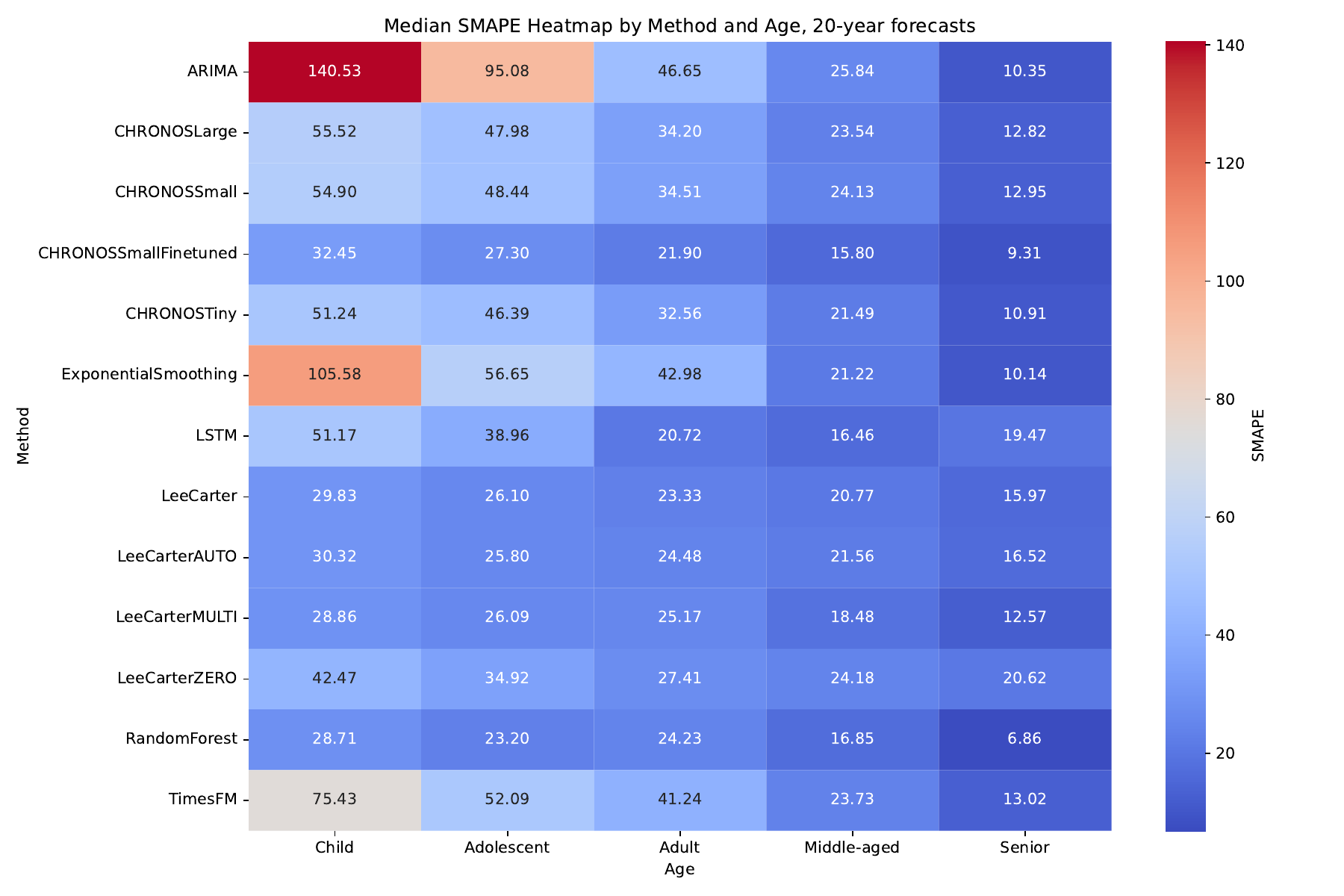}
\caption{Heatmap of median SMAPE by age categories, 20-year forecasts}
\label{fig:method_age_smape_median_20}
\end{figure}

\begin{figure}[H]
\centering
\includegraphics[width=\textwidth]{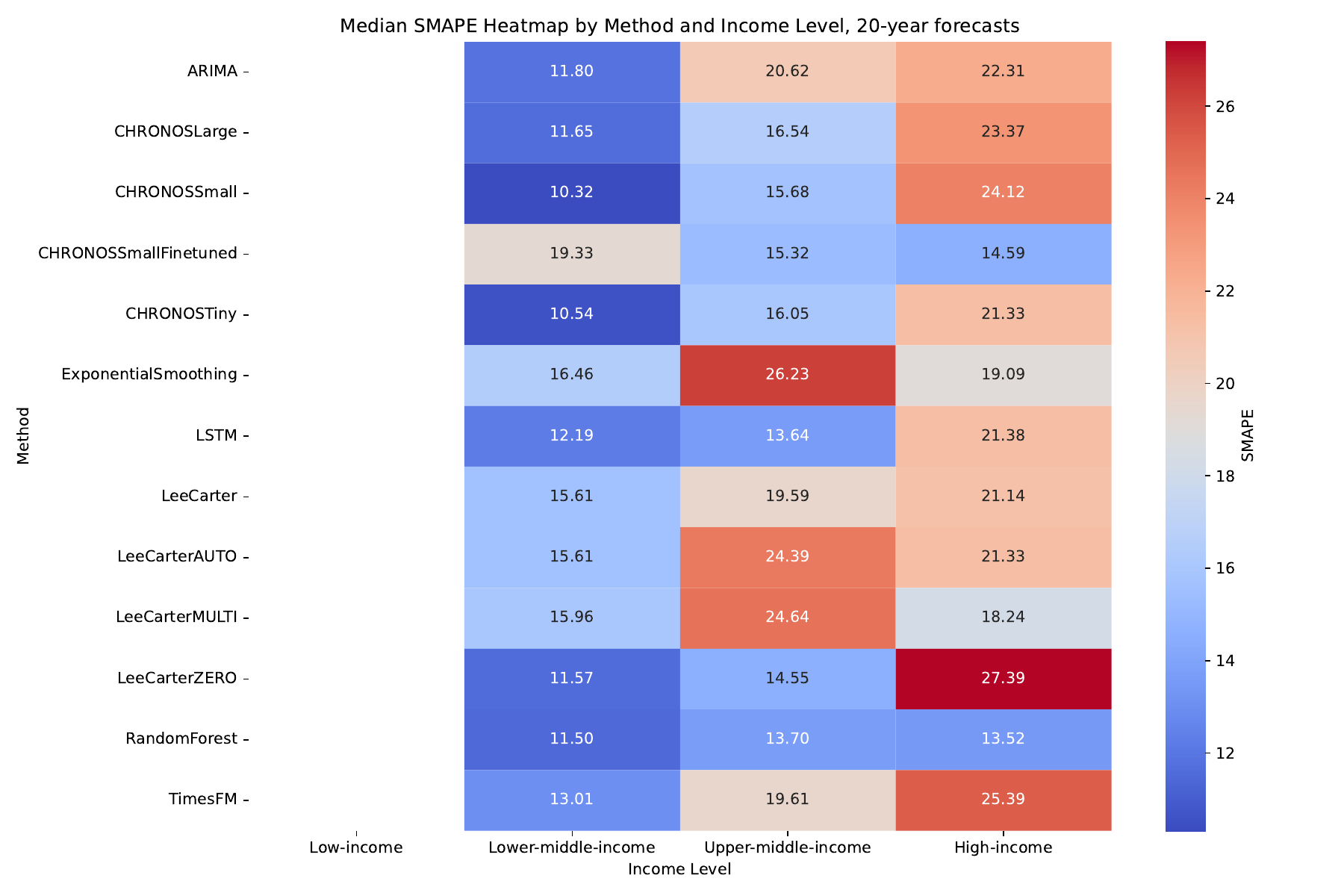}
\caption{Heatmap of median SMAPE by income categories, 20-year forecasts}
\label{fig:method_income_smape_median_20}
\end{figure}

\begin{figure}[H]
\centering
\includegraphics[width=\textwidth]{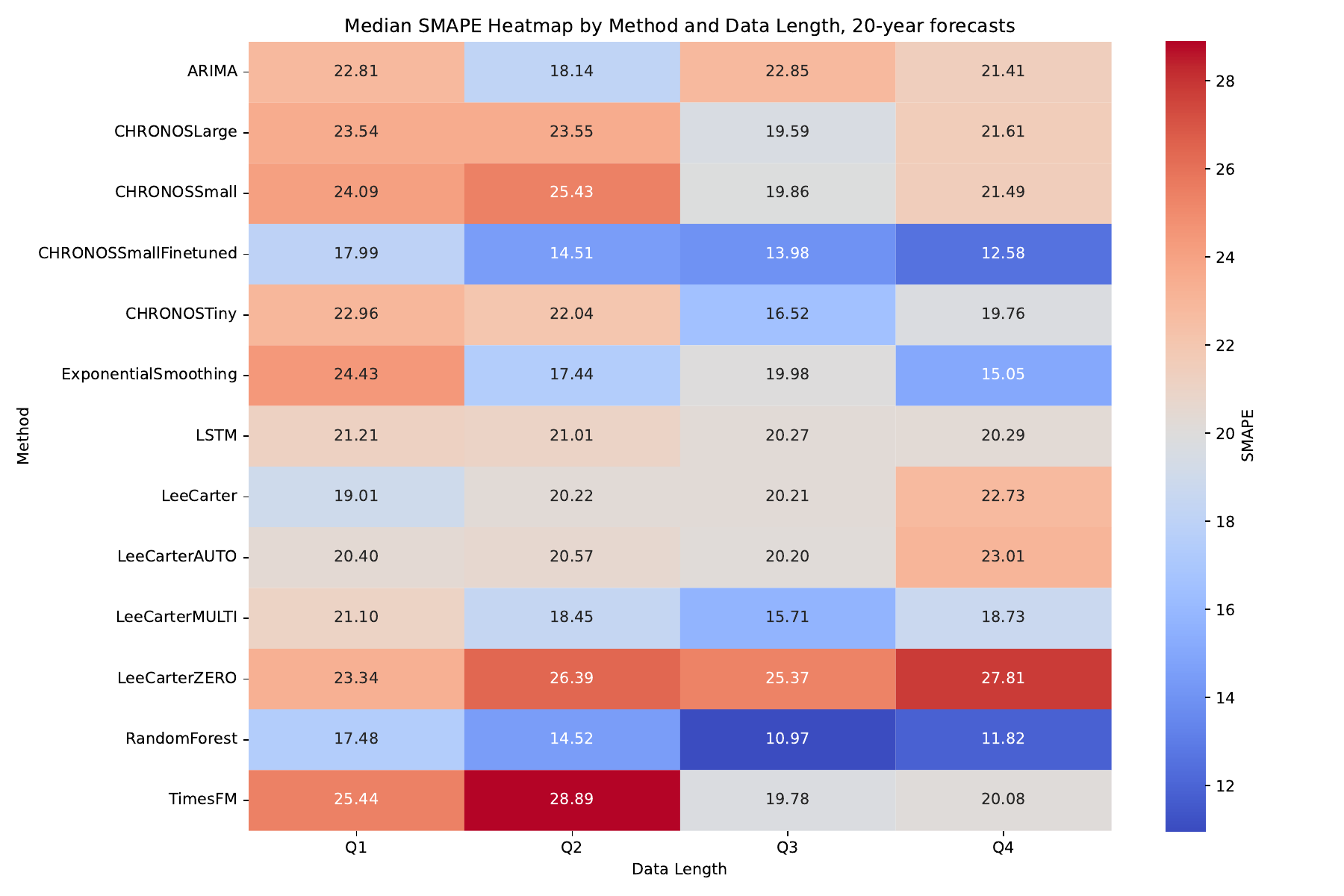}
\caption{Heatmap of median SMAPE by data length categories, 20-year forecasts}
\label{fig:method_length_smape_median_20}
\end{figure}

The median errors by age category, income category and data length category are shown in Figures \ref{fig:method_age_smape_median_20}, \ref{fig:method_income_smape_median_20} and \ref{fig:method_length_smape_median_20}, respectively. The general pre-trained models show relatively poor performance for countries with few recorded mortality rates.

\subsection{Forecasts for the Future}
\begin{figure}[H]
\centering
\includegraphics[width=\textwidth]{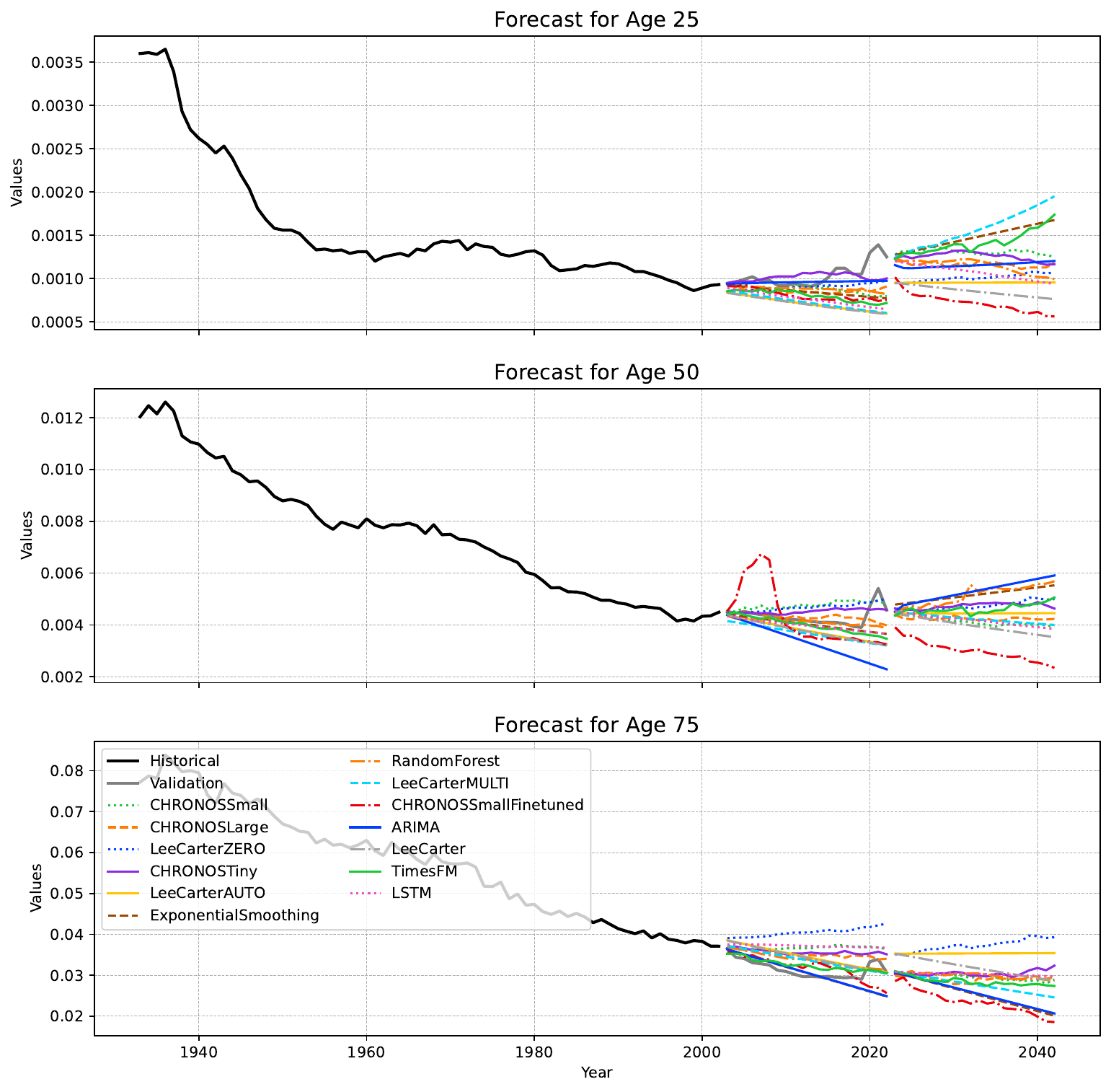}
\caption{20-year validation and future forecasts for the United States of America, for ages 25, 50 and 75}
\label{fig:forecast_USA_selected_ages_20}
\end{figure}

\begin{figure}[H]
\centering
\includegraphics[width=\textwidth]{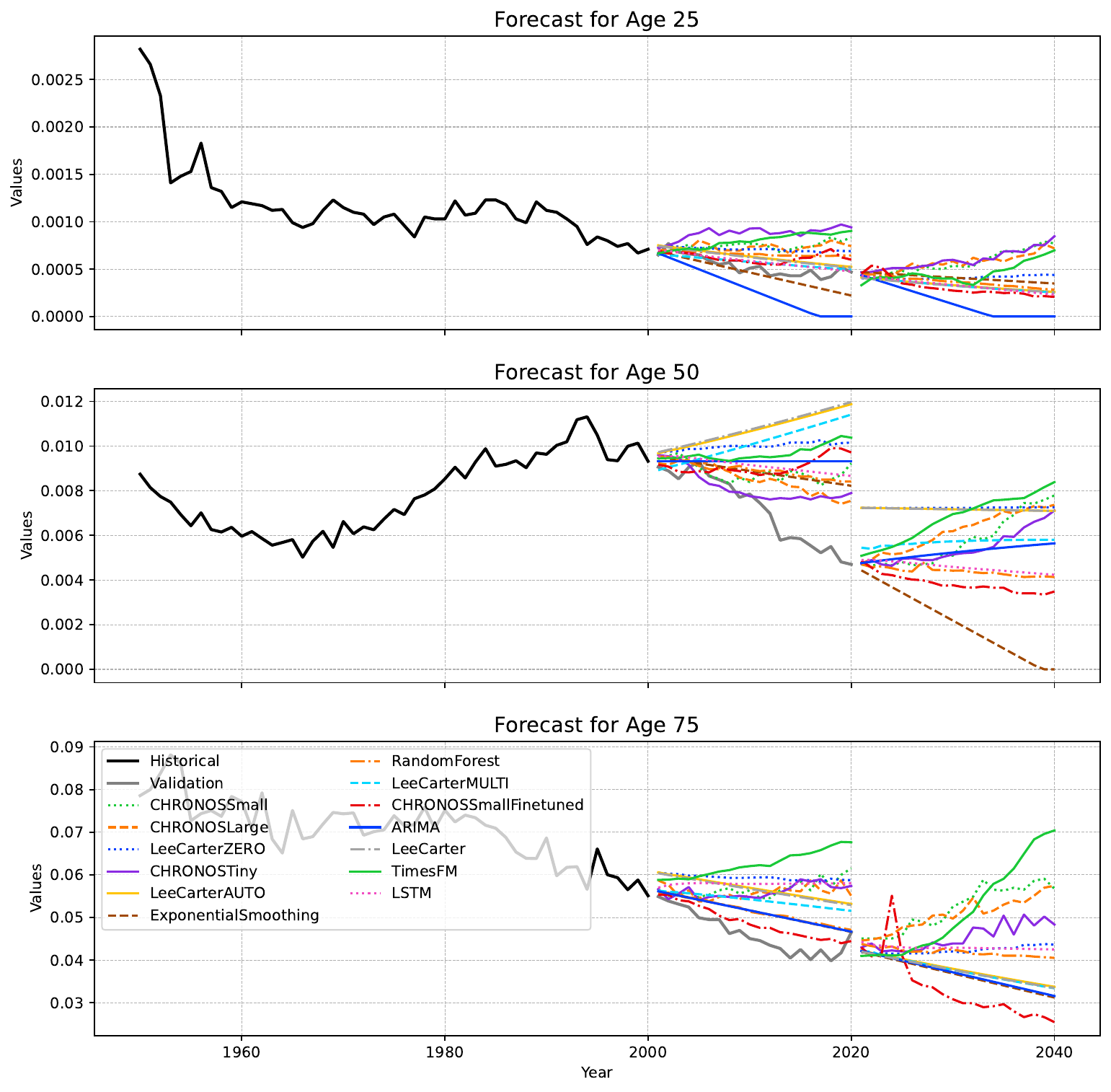}
\caption{20-year validation and future forecasts for Hungary, for ages 25, 50 and 75}
\label{fig:forecast_HUN_selected_ages_20}
\end{figure}

\begin{figure}[H]
\centering
\includegraphics[width=\textwidth]{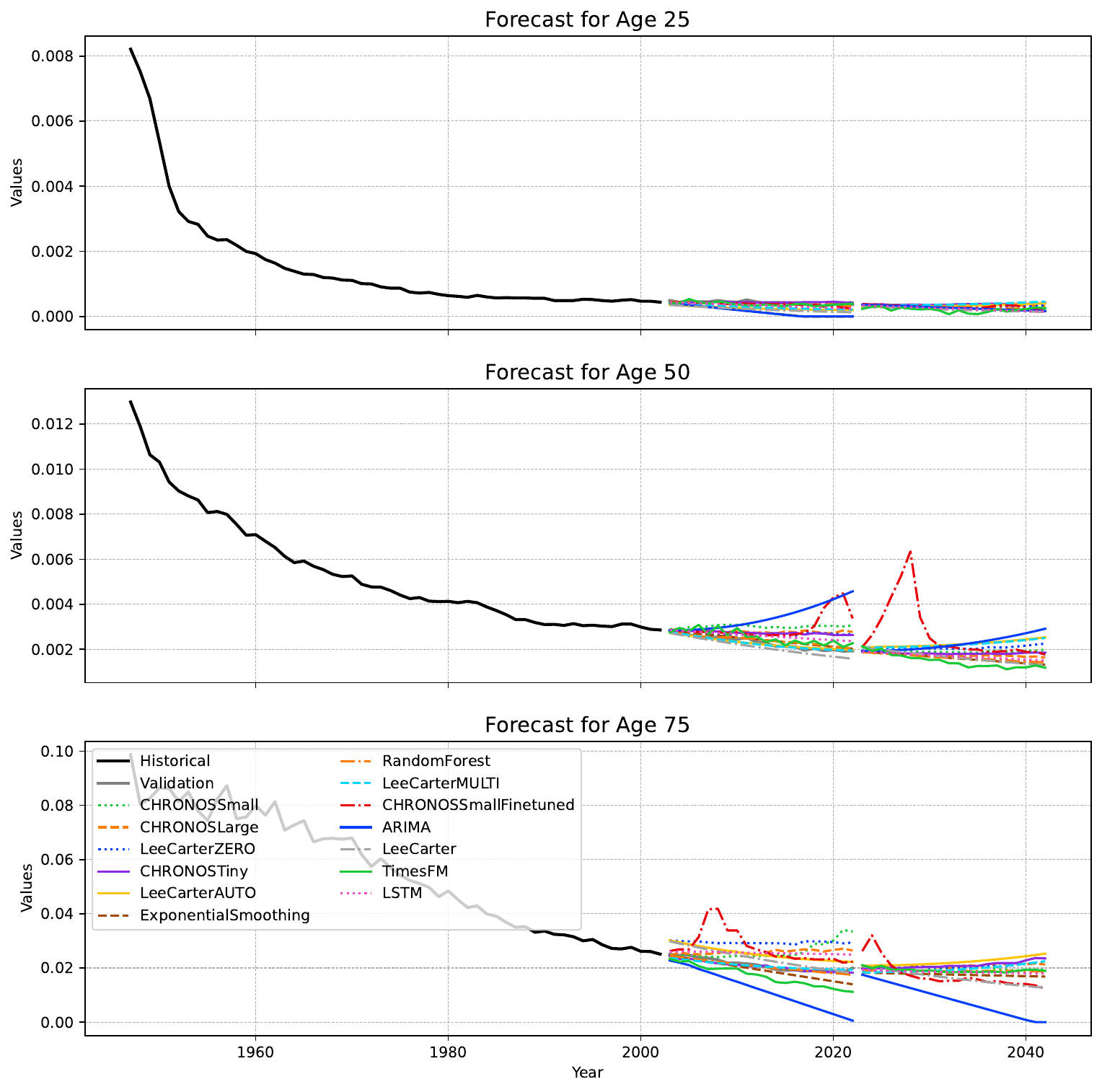}
\caption{20-year validation and future forecasts for Japan, for ages 25, 50 and 75}
\label{fig:forecast_JPN_selected_ages_20}
\end{figure}

In addition to validating the predictions on the last 5-10-20 years of data, we also made forecasts for the future, for the same time horizons. The future forecasts (together with the validation-period forecasts and the true data) for 3 arbitrarily chosen countries (United States, Hungary, Japan) are displayed in Figures \ref{fig:forecast_USA_selected_ages_20}, \ref{fig:forecast_HUN_selected_ages_20} and \ref{fig:forecast_JPN_selected_ages_20}. (Due to space constraints, only the longest-term forecasts are depicted.)

\section{Discussion}
This study explored the forecasting performance of zero-shot models, specifically TimesFM and  CHRONOS, across different time horizons, in a global context. Overall, the results highlight distinct strengths and weaknesses of zero-shot approaches compared to both machine learning models and traditional forecasting methods.
\subsection{Performance of Zero-Shot Models}
Zero-shot models exhibited considerable promise, although the performance of the two tested models differed markedly.

CHRONOS models achieved competitive median SMAPE values in short- to medium-term forecasts, often surpassing traditional models such as Lee-Carter model and ARIMA. However, they struggled in longer-term projections, particularly for populations with limited historical data, as shown by the 20-year forecasts.

TimesFM consistently lagged behind CHRONOS and other models in terms of median error. This discrepancy suggests that TimesFM’s architecture or pre-training strategy may be less suited to mortality rate prediction.
\subsection{Machine Learning vs. Traditional Models}
Machine learning models (zero-shot, pre-trained and fine-tuned) demonstrated significant advantages over traditional methods like ARIMA and Exponential Smoothing. The Random Forest trained specifically on mortality data frequently achieved the lowest SMAPE values. Traditional models occasionally rivaled machine learning methods for older age groups and in certain low-data scenarios.
\subsection{Impact of Fine-Tuning}
Fine-tuning a pre-trained CHRONOS model on mortality-specific data markedly enhanced its performance in longer-term forecasts. For the longest forecast horizon (20 years), the fine-tuned CHRONOS models achieved SMAPE reductions of 5+ percentage points compared to their zero-shot counterparts, demonstrating the value of adapting pre-trained models to domain-specific tasks. Fine-tuning allowed these models to better handle edge cases, such as mortality trends in younger age groups and countries with shorter data histories.
\subsection{Limitations}
We used only 2 pre-trained models in this study. Further models could have been used and should be used in the future, see for example ~\cite{garza2023timegpt, woo2024unified, gruver2024large, dooley2024forecastpfn, rasul2023lag}. The machine learning models that we trained on mortality data were not optimized, in the sense that no consistent hyperparameter optimization was performed. The parameters of the fine-tuning process were not optimized either.

The results were evaluated over a range of time periods (5, 10, 20 years), but in each case only the last available period was used for validation. It is possible that different results would have been obtained if earlier time periods had been considered.
\section{Conclusions}
This study assessed the zero-shot mortality rate forecasting performance of pre-trained machine learning models---specifically, neural network algorithms trained on large amounts of time series data and applied to forecast previously unseen time series.

We wanted to validate the accuracy of the projections as widely as possible, so we looked at three time periods (5, 10 and 20 years) with 111 age groups in (almost) 50 countries.

We used two zero-shot models that exhibited different levels of performance. While TimesFM was consistently among the underperforming models, CHRONOS was competitive with the top models, even with no mortality-specific fine-tuning. The zero-shot models performed less well over longer horizons. They also had difficulties forecasting the mortality rates of countries with short mortality rate histories, and forecasting the mortality rates for younger age groups. In these cases, domain-specific fine-tuning helped.

Traditional forecasting models (including the widely applied Lee-Carter model) have largely been outperformed by machine learning-based models. As a byproduct, we found that a standard machine learning model (Random Forest) trained on lagged values of the mortality rates, can produce fairly accurate forecasts.

Since zero-shot forecasting itself, although heavily researched, is still in its infancy, the results presented here are certainly encouraging for its application in the field of mortality rate forecasting.

\printbibliography

@book{box2015time,
  title={Time series analysis: forecasting and control},
  author={Box, George EP and Jenkins, Gwilym M and Reinsel, Gregory C and Ljung, Greta M},
  year={2015},
  publisher={John Wiley \& Sons}
}

@book{hyndman2018forecasting,
  title={Forecasting: principles and practice},
  author={Hyndman, RJ},
  year={2018},
  publisher={OTexts}
}

@article{hochreiter1997long,
  title={Long Short-term Memory},
  author={Hochreiter, S},
  journal={Neural Computation MIT-Press},
  year={1997}
}

@article{breiman2001random,
  title={Random forests},
  author={Breiman, Leo},
  journal={Machine learning},
  volume={45},
  pages={5--32},
  year={2001},
  publisher={Springer}
}

@article{lee1992modeling,
  title={Modeling and forecasting US mortality},
  author={Lee, Ronald D and Carter, Lawrence R},
  journal={Journal of the American statistical association},
  volume={87},
  number={419},
  pages={659--671},
  year={1992},
  publisher={Taylor \& Francis}
}

@article{booth2001age,
  title={Age-time interactions in mortality projection: Applying Lee-Carter to Australia},
  author={Booth, Heather and Maindonald, JH and Smith, Len and others},
  year={2001},
  publisher={The Australian National University}
}

@article{das2023decoder,
  title={A decoder-only foundation model for time-series forecasting},
  author={Das, Abhimanyu and Kong, Weihao and Sen, Rajat and Zhou, Yichen},
  journal={arXiv preprint arXiv:2310.10688},
  year={2023}
}

@article{nie2022time,
  title={A time series is worth 64 words: Long-term forecasting with transformers},
  author={Nie, Yuqi and Nguyen, Nam H and Sinthong, Phanwadee and Kalagnanam, Jayant},
  journal={arXiv preprint arXiv:2211.14730},
  year={2022}
}

@article{ansari2024chronos,
  title={Chronos: Learning the language of time series},
  author={Ansari, Abdul Fatir and Stella, Lorenzo and Turkmen, Caner and Zhang, Xiyuan and Mercado, Pedro and Shen, Huibin and Shchur, Oleksandr and Rangapuram, Syama Sundar and Arango, Sebastian Pineda and Kapoor, Shubham and others},
  journal={arXiv preprint arXiv:2403.07815},
  year={2024}
}

@article{radford2019language,
  title={Language models are unsupervised multitask learners},
  author={Radford, Alec and Wu, Jeffrey and Child, Rewon and Luan, David and Amodei, Dario and Sutskever, Ilya and others},
  journal={OpenAI blog},
  volume={1},
  number={8},
  pages={9},
  year={2019}
}

@article{raffel2020exploring,
  title={Exploring the limits of transfer learning with a unified text-to-text transformer},
  author={Raffel, Colin and Shazeer, Noam and Roberts, Adam and Lee, Katherine and Narang, Sharan and Matena, Michael and Zhou, Yanqi and Li, Wei and Liu, Peter J},
  journal={Journal of machine learning research},
  volume={21},
  number={140},
  pages={1--67},
  year={2020}
}

@misc{HMD,
  author       = {Human Mortality Database},
  title        = {Human Mortality Database},
  year         = {2024},
  note         = {Available at \url{https://www.mortality.org}},
  howpublished = {\url{https://www.mortality.org}},
}

@article{garza2023timegpt,
  title={TimeGPT-1},
  author={Garza, Azul and Mergenthaler-Canseco, Max},
  journal={arXiv preprint arXiv:2310.03589},
  year={2023}
}

@article{woo2024unified,
  title={Unified training of universal time series forecasting transformers},
  author={Woo, Gerald and Liu, Chenghao and Kumar, Akshat and Xiong, Caiming and Savarese, Silvio and Sahoo, Doyen},
  journal={arXiv preprint arXiv:2402.02592},
  year={2024}
}

@article{gruver2024large,
  title={Large language models are zero-shot time series forecasters},
  author={Gruver, Nate and Finzi, Marc and Qiu, Shikai and Wilson, Andrew G},
  journal={Advances in Neural Information Processing Systems},
  volume={36},
  year={2024}
}

@article{dooley2024forecastpfn,
  title={Forecastpfn: Synthetically-trained zero-shot forecasting},
  author={Dooley, Samuel and Khurana, Gurnoor Singh and Mohapatra, Chirag and Naidu, Siddartha V and White, Colin},
  journal={Advances in Neural Information Processing Systems},
  volume={36},
  year={2024}
}

@inproceedings{rasul2023lag,
  title={Lag-llama: Towards foundation models for time series forecasting},
  author={Rasul, Kashif and Ashok, Arjun and Williams, Andrew Robert and Khorasani, Arian and Adamopoulos, George and Bhagwatkar, Rishika and Bilo{\v{s}}, Marin and Ghonia, Hena and Hassen, Nadhir and Schneider, Anderson and others},
  booktitle={R0-FoMo: Robustness of Few-shot and Zero-shot Learning in Large Foundation Models},
  year={2023}
}

@article{levantesi2019application,
  title={Application of machine learning to mortality modeling and forecasting},
  author={Levantesi, Susanna and Pizzorusso, Virginia},
  journal={Risks},
  volume={7},
  number={1},
  pages={26},
  year={2019},
  publisher={MDPI}
}

@article{petnehazi2019mortality,
  title={Mortality rate forecasting: can recurrent neural networks beat the Lee-Carter model?},
  author={Petneh{\'a}zi, G{\'a}bor and G{\'a}ll, J{\'o}zsef},
  journal={arXiv preprint arXiv:1909.05501},
  year={2019}
}

@article{richman2021neural,
  title={A neural network extension of the Lee--Carter model to multiple populations},
  author={Richman, Ronald and W{\"u}thrich, Mario V},
  journal={Annals of Actuarial Science},
  volume={15},
  number={2},
  pages={346--366},
  year={2021},
  publisher={Cambridge University Press}
}

@article{nigri2019deep,
  title={A deep learning integrated Lee--Carter model},
  author={Nigri, Andrea and Levantesi, Susanna and Marino, Mario and Scognamiglio, Salvatore and Perla, Francesca},
  journal={Risks},
  volume={7},
  number={1},
  pages={33},
  year={2019},
  publisher={MDPI}
}

@article{scognamiglio2022calibrating,
  title={Calibrating the lee-carter and the poisson lee-carter models via neural networks},
  author={Scognamiglio, Salvatore},
  journal={ASTIN Bulletin: The Journal of the IAA},
  volume={52},
  number={2},
  pages={519--561},
  year={2022},
  publisher={Cambridge University Press}
}

@article{corsaro2024quantile,
  title={Quantile mortality modelling of multiple populations via neural networks},
  author={Corsaro, Stefania and Marino, Zelda and Scognamiglio, Salvatore},
  journal={Insurance: Mathematics and Economics},
  volume={116},
  pages={114--133},
  year={2024},
  publisher={Elsevier}
}

@article{wang2024time,
  title={Time-series forecasting of mortality rates using transformer},
  author={Wang, Jun and Wen, Lihong and Xiao, Lu and Wang, Chaojie},
  journal={Scandinavian Actuarial Journal},
  volume={2024},
  number={2},
  pages={109--123},
  year={2024},
  publisher={Taylor \& Francis}
}

@article{roshani2022transformer,
  title={Transformer self-attention network for forecasting mortality rates},
  author={Roshani, Amin and Izadi, Muhyiddin and Khaledi, Baha-Eldin},
  journal={Journal of the Iranian Statistical Society},
  volume={21},
  number={1},
  pages={81--103},
  year={2022},
  publisher={Iranian Statistical Society}
}

@article{makhonza2024mortality,
  title={Mortality Forecasting Using Temporal Fusion Transformers},
  author={Makhonza, Bhekamandaba and Mogodi, Nape and Mbuvha, Rendani},
  journal={Available at SSRN 4684436},
  year={2024}
}

@article{lim2021temporal,
  title={Temporal fusion transformers for interpretable multi-horizon time series forecasting},
  author={Lim, Bryan and Ar{\i}k, Sercan {\"O} and Loeff, Nicolas and Pfister, Tomas},
  journal={International Journal of Forecasting},
  volume={37},
  number={4},
  pages={1748--1764},
  year={2021},
  publisher={Elsevier}
}

@article{vaswani2017attention,
  title={Attention is all you need},
  author={Vaswani, Ashish and Shazeer, Noam and Parmar, Niki and Uszkoreit, Jakob and Jones, Llion and Gomez, Aidan N and Kaiser, {\L}ukasz and Polosukhin, Illia},
  journal={Advances in neural information processing systems},
  volume={30},
  year={2017}
}

@article{radford2018improving,
  title={Improving language understanding by generative pre-training},
  author={Radford, Alec and Narasimhan, Karthik and Salimans, Tim and Sutskever, Ilya and others},
  year={2018},
  publisher={San Francisco, CA, USA}
}

@article{hainaut2018neural,
  title={A neural-network analyzer for mortality forecast},
  author={Hainaut, Donatien},
  journal={ASTIN Bulletin: The Journal of the IAA},
  volume={48},
  number={2},
  pages={481--508},
  year={2018},
  publisher={Cambridge University Press}
}

@article{schnurch2022point,
  title={Point and interval forecasts of death rates using neural networks},
  author={Schn{\"u}rch, Simon and Korn, Ralf},
  journal={ASTIN Bulletin: The Journal of the IAA},
  volume={52},
  number={1},
  pages={333--360},
  year={2022},
  publisher={Cambridge University Press}
}

@article{hyndman2008automatic,
  title={Automatic time series forecasting: the forecast package for R},
  author={Hyndman, Rob J and Khandakar, Yeasmin},
  journal={Journal of statistical software},
  volume={27},
  pages={1--22},
  year={2008}
}

@MISC {pmdarima,
  author = {Taylor G. Smith and others},
  title  = {{pmdarima}: ARIMA estimators for {Python}},
  year   = {2017--},
  url    = "http://www.alkaline-ml.com/pmdarima",
  %note   = {[Online; accessed 2025-03-11]}
}

@inproceedings{seabold2010statsmodels,
  title={statsmodels: Econometric and statistical modeling with python},
  author={Seabold, Skipper and Perktold, Josef},
  booktitle={9th Python in Science Conference},
  year={2010},
}

@misc{chollet2015keras,
  title={Keras},
  author={Chollet, Fran\c{c}ois and others},
  year={2015},
  howpublished={\url{https://keras.io}},
}

@article{scikit-learn,
  title={Scikit-learn: Machine Learning in {P}ython},
  author={Pedregosa, F. and Varoquaux, G. and Gramfort, A. and Michel, V.
          and Thirion, B. and Grisel, O. and Blondel, M. and Prettenhofer, P.
          and Weiss, R. and Dubourg, V. and Vanderplas, J. and Passos, A. and
          Cournapeau, D. and Brucher, M. and Perrot, M. and Duchesnay, E.},
  journal={Journal of Machine Learning Research},
  volume={12},
  pages={2825--2830},
  year={2011}
}

@article{baran2007forecasting,
  title={Forecasting Hungarian mortality rates using the Lee-Carter method},
  author={Baran, S{\'a}ndor and G{\'a}ll, J{\'o}zsef and Isp{\'a}ny, M{\'a}rton and Pap, Gyula},
  journal={Acta Oeconomica},
  volume={57},
  number={1},
  pages={21--34},
  year={2007},
  publisher={Akad{\'e}miai Kiad{\'o}}
}

\end{document}